\documentclass[letterpaper,10pt,conference]{ieeeconf}

\IEEEoverridecommandlockouts
\usepackage{amsmath}
\usepackage{amssymb}
\usepackage{array}
\usepackage{booktabs}
\usepackage{cite}
\usepackage{graphicx}
\usepackage{microtype}
\usepackage{xcolor}
\usepackage{url}
\usepackage[hidelinks]{hyperref}
\hypersetup{
  pdftitle={PopNavShift: Stress-Testing Social Navigation under Behavioral Population Shift},
  pdfauthor={Kaizhen Tan, Diyu Zheng, Tim Guangyu Wu, ChengHe Guan}
}

\title{\LARGE \bf
PopNavShift: Stress-Testing Social Navigation\\
under Behavioral Population Shift
}

\definecolor{projectlink}{HTML}{225E66}
\author{Kaizhen Tan$^{1,2}$, Diyu Zheng$^{1,2}$, Tim Guangyu Wu$^{1,2}$, and ChengHe Guan$^{2,3}$\\[0.7ex]
{\small $^{1}$Robert F. Wagner Graduate School of Public Service, New York University, New York, NY, USA}\\
{\small $^{2}$Shanghai Key Laboratory of Urban Design and Urban Science, NYU Shanghai, Shanghai, China}\\
{\small $^{3}$Division of Arts and Sciences, NYU Shanghai, Shanghai, China}}

\begin{document}
\maketitle
\thispagestyle{empty}
\pagestyle{empty}

\begin{abstract}
Social-navigation algorithms are often evaluated under a fixed pedestrian-behavior distribution, despite substantial variation in pedestrian responses to robots across individuals and social contexts. We introduce PopNavShift, a matched simulation framework for stress-testing social-navigation strategies under pedestrian population shifts. PopNavShift constructs population-conditioned pedestrian motion profiles by prompting Gemini 3.7 Flash with 600 synthetic persona records from MatrAIx Persona 1M and deterministically mapping the responses into bounded motion parameters. It then compares three representative navigation strategies, reactive avoidance, early yielding, and reciprocal collision avoidance, across eight population conditions and 7,488 matched robot runs. In a matched intervention on the same 202 personas, changing only time pressure reverses 8.6\% of controller rankings based on robot travel time, but 22.4\% based on mean pedestrian delay and 23.9\% based on worst-decile delay. Across population conditions, this sensitivity is greater for pedestrian burden than for robot travel time and increases in spatially constrained settings; the same qualitative pattern persists under a second pedestrian dynamics model. These findings support evaluating navigation strategies across behavioral populations using both robot performance and pedestrian burden.
\end{abstract}

\section{Introduction}

As robots increasingly operate in human-shared spaces, their navigation must account not only for task efficiency but also for how nearby pedestrians respond to them. These responses can vary substantially across people and situations, with differences in how much clearance pedestrians maintain, when they yield, whether they accept small gaps, and how quickly they move around an approaching robot. Human--robot proxemics and social-navigation studies show that these behaviors depend on robot motion, orientation, gaze, encounter context, and social conventions \cite{Takayama2009,Mumm2011,Xu2025,ikeda2025overlap}. In simulation-based evaluation, pedestrian behavior is often represented through a selected model or behavioral distribution, making the represented population part of the evaluation setup alongside the physical scenario.

Existing work accounts for variation in pedestrian behavior. Social-navigation methods incorporate pedestrian emotion, individual preferences, and social judgments into planning \cite{narayanan2023ewarenet,choi2020fast,wang2025navidiff}, while platforms such as SOCIALGYM and models such as MAC-ID support configurable or heterogeneous pedestrian behaviors \cite{holtz2022socialgym,Chung2024}. However, these approaches do not directly test whether the relative performance of the same navigation strategies remains stable when the represented pedestrian population changes. Navigation strategies can distribute interaction adjustment differently between the robot and nearby pedestrians. Changes in pedestrian yielding, clearance preference, or gap acceptance may therefore alter their relative performance even when geometry, density, and initial conditions remain fixed. Although current evaluation guidelines emphasize human, environmental, and task context \cite{francis2025principles}, the stability of navigation-strategy rankings across behavioral population shifts has received limited direct study.

\begin{figure*}[t]
    \centering
    \includegraphics[width=0.95\textwidth]{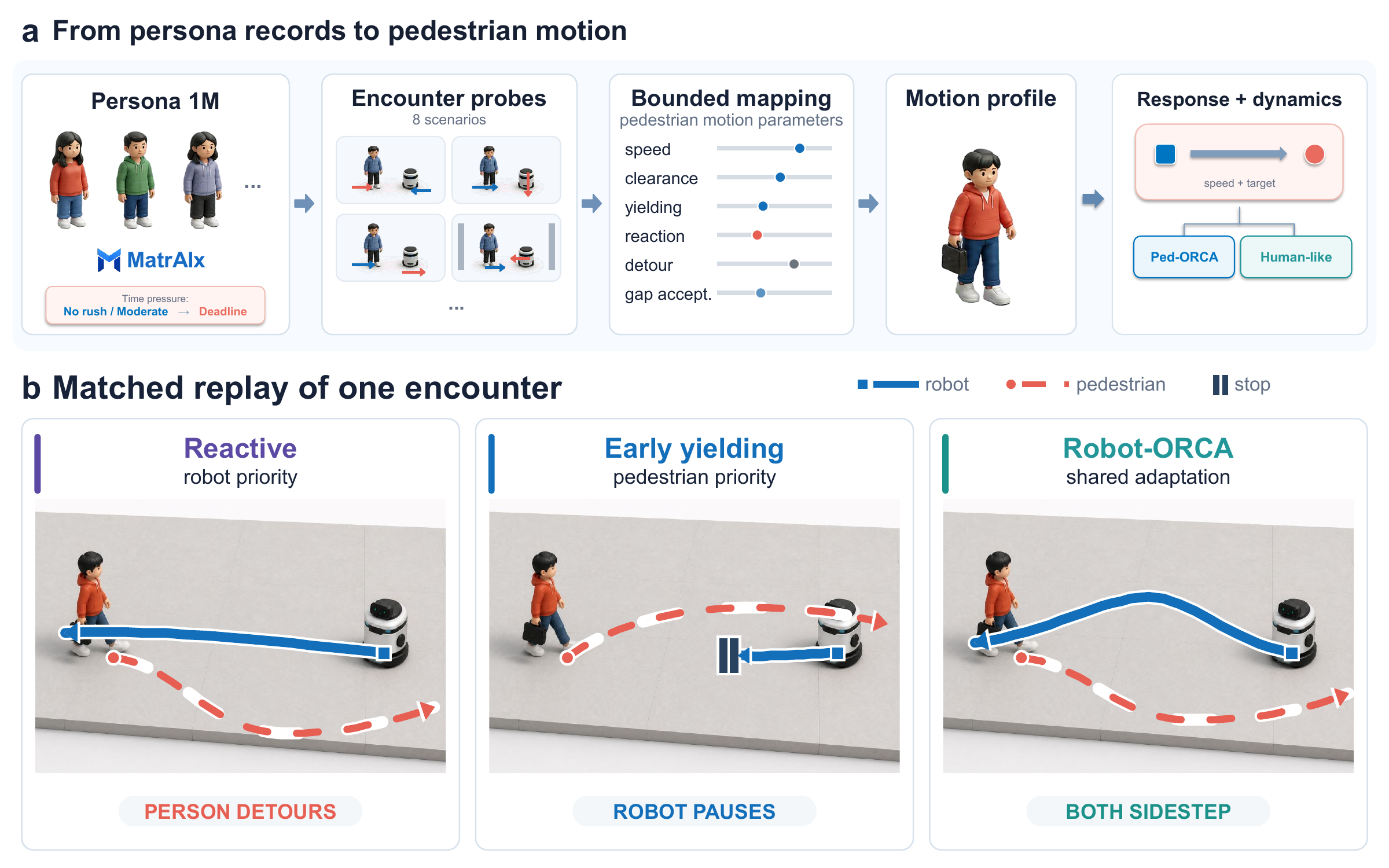}
    \caption{PopNavShift evaluates social navigation strategies under behavioral
    population shift. (a) Persona records are converted through
    standardized robot encounter probes and a deterministic bounded mapping into
    pedestrian motion profiles. The matched intervention changes time pressure
    for the same persona records before repeating the probe-to-motion pipeline.
    (b) Identical physical encounters are replayed under reactive geometric,
    early yielding, and Robot-ORCA controllers. Ped-ORCA is the primary pedestrian
    dynamics backend; navground Human-like behavior is used for robustness
    analysis.}
    \label{fig:framework}
\end{figure*}

We therefore introduce \textbf{PopNavShift}, a matched evaluation framework that holds physical episodes and navigation strategies fixed while varying the represented pedestrian population (Fig.~\ref{fig:framework}). This work makes three contributions. First, we establish behavioral population shift as an explicit dimension of social-navigation evaluation, making robustness to population change a direct evaluation target. Second, we operationalize population robustness through matched population replay and a same-persona time-pressure intervention, assessing changes in navigation-strategy rankings based on robot travel time and pedestrian burden, including mean and worst-decile added delay relative to no-robot baselines. Third, we examine how population sensitivity varies with spatial constraints and whether ranking sensitivity persists across alternative response-to-motion mappings, pedestrian dynamics models, and language models.

\section{Related Work}

\subsection{Human Variation in Social Navigation}

Human--robot interaction studies document variation in proxemics, perceived
safety, passing conventions, and preferred paths in shared environments
\cite{francis2025principles,ikeda2025overlap}. Robot approach direction,
orientation, gaze, and motion can affect the spacing people maintain during
encounters \cite{Takayama2009,Mumm2011,Xu2025}. These findings show that
pedestrian spatial behavior depends on encounter geometry and interaction cues.

Behavior also varies across populations and situational contexts. Eresha et al.\
report cultural differences in proxemic behavior around humanoid robots
\cite{Eresha2013}, while Xiao et al.\ compare collision-avoidance behavior among
participants in the United States and Israel \cite{Xiao2024}. Time pressure can
affect crossing speed and gap acceptance \cite{tian2022pressure}, motivating its
use as the controlled contextual variable in our matched intervention.

Other work incorporates individual states and preferences into robot navigation.
EWareNet uses inferred pedestrian emotion to adapt intent prediction and spatial
profiles \cite{narayanan2023ewarenet}; Choi et al.\ infer user-specific
navigation preferences \cite{choi2020fast}; and NaviDIFF incorporates human
preferences and social principles into navigation behavior
\cite{wang2025navidiff}. These methods adapt robot behavior to individual users,
preferences, or inferred states. PopNavShift keeps the navigation strategies
fixed and examines whether their relative rankings remain stable as the
represented pedestrian population changes.

\subsection{Pedestrian Diversity in Simulation}

To represent variation in pedestrian behavior, social-navigation simulation
relies on motion models that specify how people move and interact around robots.
Classical approaches such as the Social Force model and reciprocal collision
avoidance generate pedestrian motion from explicit interaction rules
\cite{helbing1995social,berg2011orca}, while other approaches learn crowd motion
or interaction behavior from observed trajectories. These models provide the
basis for simulating pedestrian responses, with behavioral variation typically
represented through shared model parameters or predefined behavior settings.

Recent simulation systems make this variation more explicit. SOCIALGYM exposes
configurable human parameters for social-navigation experiments
\cite{holtz2022socialgym}, while SEAN 2.0 supports multiple methods for specifying
pedestrian motion and generating varied social situations \cite{Tsoi2022}.
MAC-ID models individual differences through a local coordination factor that
varies pedestrian behavior from more competitive to more cooperative
interaction, and its BSON benchmark uses these behaviors to construct diverse
social environments for navigation evaluation \cite{Chung2024}. These systems
show that pedestrian behavior can be varied systematically within simulation,
providing a basis for evaluating navigation methods across different behavioral
conditions.

Recorded human-motion datasets provide another source for grounding simulated pedestrian behavior. SCAND, TH\"OR, and TH\"OR-MAGNI capture pedestrian motion and human--robot encounters in real environments \cite{karnan2022scand,rudenko2020thor,schreiter2025thormagni}. Such datasets can inform plausible motion patterns and physical parameter ranges. Our matched population-shift design additionally requires structured contextual attributes that can be modified while keeping the underlying persona fixed. The recorded datasets above do not provide this intervention structure. 

\subsection{Population Robustness in Social Navigation} 

Alongside work on modeling pedestrian diversity, prior research has developed standardized protocols and benchmarks for evaluating social-navigation methods. SocNavBench replays recorded pedestrian trajectories under a common evaluation protocol \cite{biswas2022socnavbench}, while SocNav1 provides socially judged navigation situations for evaluating social conventions \cite{manso2020socnav1}. Recent evaluation guidelines further organize metrics, scenarios, and human, environmental, and task contexts for social-navigation assessment \cite{francis2025principles}.

BSON is especially relevant because it evaluates navigation methods in environments with diverse pedestrian interaction styles \cite{Chung2024}. Built on MAC-ID, it uses pedestrian coordination styles ranging from more competitive to more cooperative behavior to construct diverse social environments. PopNavShift focuses on whether the relative performance of the same navigation strategies remains stable as the represented pedestrian population changes while the physical encounter is held fixed. We evaluate this through matched population replay, comparing changes in controller rankings and pedestrian burden across population conditions.

\section{Method}

\subsection{Framework Overview}

To operationalize the matched population-replay design, PopNavShift separates the construction of behavioral populations from pedestrian dynamics and robot control (Fig.~\ref{fig:framework}). Eight fixed robot-encounter probes, each describing a standardized navigation situation, elicit structured responses from each persona, which a deterministic mapping converts into seven bounded motion controls. The resulting motion profiles are then organized into eight population conditions for matched evaluation using a shared pedestrian response layer and continuous dynamics model.

\subsection{Persona Records and Standardized Probes}

We sample 600 records from the public MatrAIx Persona 1M release
\cite{li2026matraix,matraix2026data}. The prompt includes available fields related
to time pressure, safety and risk preferences, decision style, trust in
technology, patience, accessibility needs, and mobility. Demographic attributes
are excluded from the prompt and mapping.

Gemini 3.7 Flash receives each record and answers eight fixed robot encounter
probes. For each encounter, it selects one action from \emph{continue},
\emph{slow}, \emph{stop}, \emph{yield}, \emph{pass}, or \emph{detour}. The probes
vary relative approach direction, available space, crowding or obstruction, and
robot behavior. Together, they cover situations in which clearance, yielding,
speed adjustment, and detouring can affect the interaction.

After selecting the encounter actions, the model returns an ordered response block
covering preferred clearance, willingness to wait, yielding, small-gap
acceptance, detouring, robot trust, and urgency on five-level scales. The complete
structured response is cached for each persona and reused across simulation
scenarios. The probes form a response-elicitation battery; they are
separate from the physical scenarios used in the simulation experiments.

\subsection{Mapping from Responses to Motion}

The probe responses are converted into continuous pedestrian controls using a
deterministic bounded mapping. The five ordered response levels are encoded as
$\{0,0.25,0.5,0.75,1\}$. For persona $i$, the motion profile is
\[
\boldsymbol{\theta}_i=(v_i,d_i,a_i,y_i,\tau_i,L_i,g_i),
\]
where the components denote preferred speed, robot clearance, avoidance
strength, yielding propensity, reaction delay, detour threshold, and gap
acceptance, respectively. These are operational simulator controls rather than
estimated psychological traits.

Let $q_i^c$, $q_i^w$, $q_i^y$, $q_i^g$, $q_i^d$, $q_i^t$, and $q_i^u$
denote the normalized responses for clearance, waiting, yielding, small-gap
acceptance, detouring, robot trust, and urgency. Mapping A also summarizes the
categorical encounter actions using fixed ordered scores, whose mean is
$q_i^a$. Its caution score is
\begin{equation}
C_i=0.34q_i^c+0.22(1-q_i^g)+0.22q_i^y+0.22q_i^a.
\end{equation}

The normalized controls combine these responses according to their
intended simulation roles: urgency affects preferred speed and yielding;
clearance and gap acceptance use their corresponding responses directly;
avoidance combines caution with robot trust; reaction delay combines urgency
and trust; and detour tendency combines the reported detour response with
caution. All coefficients are fixed before simulation and are not fitted to
observed trajectories.

Each normalized control $z_i^k$ is clipped to $[0,1]$ and mapped to its physical
range $[\ell_k,h_k]$ by
\begin{equation}
\theta_i^k=\ell_k+z_i^k(h_k-\ell_k).
\end{equation}

The physical ranges are 0.60--1.80~m/s for preferred speed, 0.30--1.20~m for
robot clearance, 0.60--2.20 for avoidance strength, 0.15--1.20~s for reaction
delay, and 1.0--4.0~m for the detour threshold. Yielding and gap acceptance remain
in $[0,1]$. Mobility-related fields apply ordered speed caps after mapping. The
resulting speed and spacing ranges overlap values reported in TH\"OR and
related datasets \cite{rudenko2020thor}.

We additionally test three alternative mappings from responses to motion. Mappings B
and C modify the weighting and action-response construction, while Mapping D
uses separable response dimensions without weighted mixtures. All four mappings
retain the same physical ranges and mobility caps. Mapping A is used in the main
experiment and Mappings B--D are used for sensitivity analysis.

\subsection{Population Conditions and Matched Intervention}

\begin{figure*}[t]
    \centering
    \includegraphics[width=0.96\textwidth]{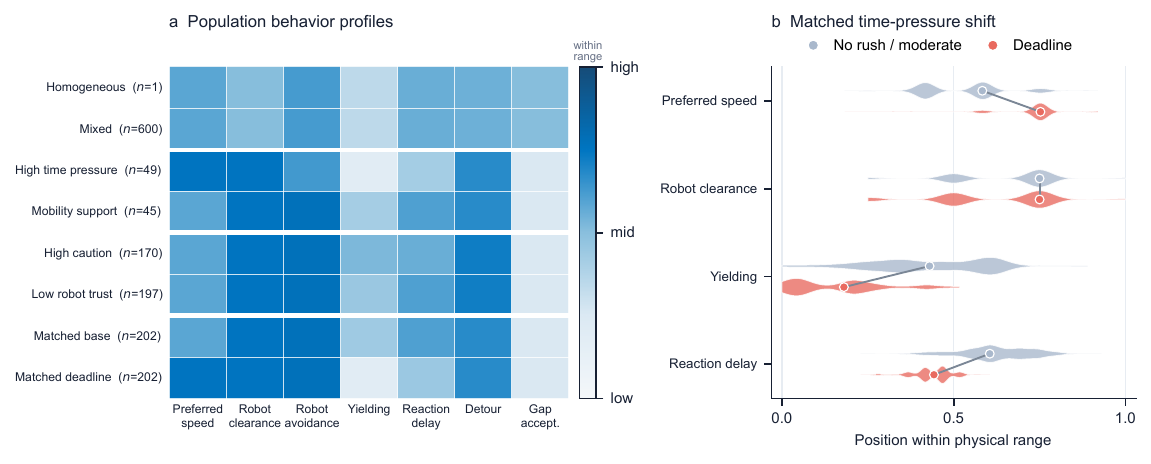}
    \caption{Pedestrian motion profiles across population conditions.
    (a) Median position within each parameter's physical range; labels report
    cohort size. White separators distinguish reference conditions, source-field
    cohorts, probe-defined cohorts, and the matched pair. (b) Profile
    distributions for the same personas before and after the matched change
    in time pressure from \emph{No rush}/\emph{Moderate} to \emph{Deadline}.}
    \label{fig:behavior-profiles}
\end{figure*}

Table~\ref{tab:populations} defines the eight population conditions used in the
experiments. These selected behavioral contrasts can overlap in persona
membership. Figure~\ref{fig:behavior-profiles}(a) summarizes the resulting
distributions over the motion parameters.

\begin{table}[t]
\caption{Behavioral Population Conditions and Cohort Sizes}
\label{tab:populations}
\centering
\footnotesize
\setlength{\tabcolsep}{3.2pt}
\begin{tabular}{@{}>{\raggedright\arraybackslash}p{0.27\columnwidth}>{\raggedright\arraybackslash}p{0.61\columnwidth}r@{}}
\toprule
Condition & Construction & $n$ \\
\midrule
Homogeneous & Overall median parameter vector & 1 \\
Mixed & All mapped records & 600 \\
High time pressure & Deadline or Emergency source field & 49 \\
Mobility support & Explicit mobility or motor-support field & 45 \\
High caution & Upper quartile of clearance/yield/gap score & 170 \\
Low robot trust & Low or very-low probe response & 197 \\
Matched base & Original no-rush/moderate records & 202 \\
Matched deadline & Same IDs; time pressure set to Deadline & 202 \\
\bottomrule
\end{tabular}
\end{table}

The matched time-pressure intervention provides the main controlled population
shift. It contains the same 202 persona records in both conditions: 113
originally have \emph{No rush} and 89 have \emph{Moderate} time pressure. The
matched-base condition uses the original records. For the matched-deadline
condition, only the time-pressure field is changed to \emph{Deadline}, after
which the same encounter probes and mapping from responses to motion are rerun.
Persona identity and all other source fields remain unchanged.

Time pressure is used as a mutable contextual variable because controlled
pedestrian studies have found that imposed time pressure can change crossing
speed and gap acceptance \cite{tian2022pressure}. This evidence motivates the
intervention; the observed effect sizes are not used to calibrate the persona
responses or motion parameters. Figure~\ref{fig:behavior-profiles}(b) shows how the
resulting motion-profile distributions change for the same personas.

For episodes containing multiple pedestrians, profiles are sampled with
replacement using fixed episode-specific assignments. Matched-base and
matched-deadline runs use the same persona at every pedestrian slot. Any
difference between the two conditions therefore originates from responses
changed by the time-pressure intervention rather than from different sampled
identities.

\subsection{Pedestrian Response and Dynamics}

Each motion profile enters simulation through a shared robot-conditioned
pedestrian response layer. The response activates after the robot enters
sensing range and the profile-specific reaction delay has elapsed. Both agents
are projected over a four-second horizon using the effective clearance
\[
d_i^{\mathrm{eff}}=d_i(1-0.35g_i).
\]
A conflict is predicted when the closest projected center distance falls below
the sum of the two agent radii and $d_i^{\mathrm{eff}}$.

During a predicted conflict, yielding propensity reduces the pedestrian's
preferred speed as the robot approaches the detour threshold $L_i$. When the
robot is nearly stationary, this reduction is relaxed to allow the pedestrian
to pass. Inside $L_i$, the response layer places a lateral waypoint 1.5~m ahead,
with offset determined by effective clearance and avoidance strength. Each
episode fixes a passing side and switches sides if an obstacle blocks the
preferred route.

Navground provides the continuous two-dimensional simulation \cite{navground}.
The primary pedestrian backend, Ped-ORCA, uses omnidirectional Optimal Reciprocal
Collision Avoidance (ORCA) with a 4.5~m spatial horizon \cite{berg2011orca}.
A robustness experiment uses
navground's Human-like (HL) behavior, which responds to visual collision cues
within a 5~m horizon \cite{guzzi2013humanfriendly}. Both backends receive the same
preferred speeds and targets from the shared response layer. The simulation
step is 0.1~s and the episode time limit is 45~s.

\subsection{Robot Controllers}

The robot is modeled as a differential-drive disk with radius 0.35~m. All
controllers share the same goal, footprint, dynamic limits, local observations,
and collision geometry.

The \emph{reactive geometric} controller samples feasible wheel commands over a
2~s horizon with a 0.04~m safety buffer and adjusts its trajectory when a conflict
becomes imminent. The \emph{early yielding} controller detects conflicts over a
4.5~s horizon, moves toward a clear pull-aside pose with a 0.16~m buffer, and
waits while the pedestrian crosses the conflict region. The \emph{Robot-ORCA}
controller uses reciprocal collision avoidance with a 5~m spatial horizon and a
0.12~m safety margin under navground's two-wheel kinematics.

The controllers therefore differ primarily in how interaction adjustment
is allocated. Reactive control postpones robot-side adjustment, early yielding
gives greater priority to the pedestrian, and Robot-ORCA uses reciprocal
adaptation. Ped-ORCA denotes the pedestrian dynamics backend; Robot-ORCA denotes
the reciprocal robot controller.

\section{Experimental Design}

\subsection{Matched Evaluation Grid}

We evaluate PopNavShift on five types of physical encounters: a bidirectional
sidewalk, a head-on encounter, a crossing, a bottleneck, and a static
obstruction. The scenario grid varies available width (1.8, 2.4, 3.0, and
4.2~m), pedestrian density (0.06, 0.12, and 0.20 persons/m$^2$), and robot
speed (0.60 and 1.00~m/s). Head-on encounters contain one pedestrian and use
the lowest density, yielding 104 valid physical cells rather than a full
Cartesian product.

Each physical cell contains three fixed variants with different pedestrian
start positions and walking directions, giving 312 episodes in total. Every
episode is replayed with all eight population conditions and all three robot
controllers. For each population--episode combination, an additional run
without a robot provides the pedestrian baseline. The matched design keeps
physical geometry, episode initial conditions, and controller definitions
unchanged when the represented pedestrian population changes.

\subsection{Metrics and Ranking Stability}

Robot-centered outcomes include success, travel time, path length, stopping
time, pedestrian and world collisions, and minimum pedestrian clearance.
Pedestrian-centered outcomes include travel time, walking distance, stopping
time, detour distance, and minimum robot clearance.

Our main pedestrian-burden measure is added travel time. For pedestrian $j$,
\[
\Delta T_j =
T_j^{\mathrm{robot}} - T_j^{\mathrm{no\ robot}},
\]
where the run without a robot uses the same physical episode and pedestrian
assignment. Added distance is defined analogously. A paired delay is available
when the pedestrian completes both runs.

A physical cell fixes scenario type, width, density, and robot speed and pools
its three episode variants. Mean pedestrian delay is computed over all
available paired records in the cell. To capture concentrated burden,
worst-decile delay is the mean of the largest $\lceil 0.1n \rceil$ paired
delays. Failed or timed-out robot runs retain the 45~s episode limit as robot
travel time.

For each physical cell and metric, we compare every pair of controllers
separately under each population. A \emph{ranking reversal} occurs when a
non-tied controller ordering under one population changes sign under another.
We use tie tolerances of 0.02~s for pedestrian delay and 0.05~s for robot
travel time. We additionally report a completed-cell analysis restricted to
cells in which all three controllers finish every episode variant under both
populations being compared.

To quantify the magnitude of population effects independently of controller
ordering, we define \emph{population sensitivity} as the range of a metric
across population conditions for a fixed controller and physical cell.

\subsection{Sensitivity Analyses}

We test whether the observed population dependence is specific to the
mapping from responses to motion or pedestrian dynamics model. Mapping sensitivity
is evaluated on twelve representative episodes spanning narrow and wide
sidewalks, bottlenecks, and static obstructions. Mappings A--D use the same
cached responses, population membership, and episode assignments. In addition
to ranking-reversal rates, we measure agreement with Mapping A on eligible
non-tied controller comparisons.

We also perturb the structure of the pedestrian profiles. One analysis
reassigns complete motion-profile vectors across persona identities before
reconstructing five shared populations. A second independently shuffles each
motion-parameter column within a population, preserving marginal parameter
distributions while breaking their cross-parameter association. Separate runs
isolate the contributions of time-pressure, caution, and trust response blocks.

Finally, we repeat a subset of the evaluation with navground's Human-like
pedestrian dynamics. This comparison uses twelve obstacle-free sidewalk,
head-on, and crossing episodes and four shared populations: homogeneous,
mixed, matched base, and matched deadline. Physical scenes, persona assignments,
response rules, and robot controllers remain matched between Ped-ORCA and
Human-like dynamics. Backend comparisons use the intersection of cells
completed by all three controllers under both models, after which the same
metric-specific tie tolerances are applied separately to each backend.

\section{Results}

\subsection{Matched Time-Pressure Shift}

The main experiment contains 7,488 robot runs and 2,496 matched runs without a robot, with paired pedestrian delay available for nearly all records.
Changing time pressure from \emph{No rush}/\emph{Moderate} to
\emph{Deadline} alters at least one categorical response for 201 of the
202 matched personas.

\begin{figure*}[t]
    \centering
    \includegraphics[width=0.90\textwidth]{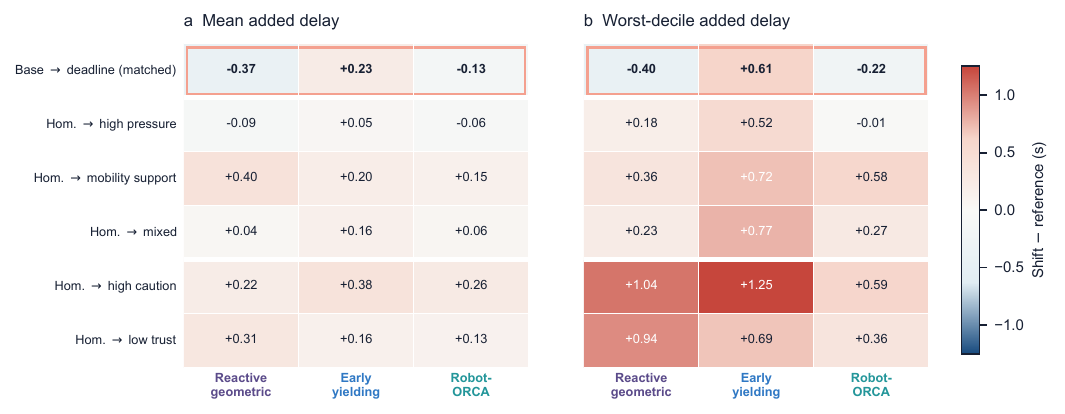}
    \caption{Changes in pedestrian delay under behavioral population shift.
    Entries show shifted minus reference delay over the same 104 physical
    cells. The outlined row shows the matched time-pressure intervention on the
    same personas. White separators distinguish the matched intervention,
    source-field cohorts, the mixed population, and probe-defined cohorts.}
    \label{fig:population-shifts}
\end{figure*}

This shift changes controller comparisons more often for pedestrian burden
than for robot travel time. Mean-delay orderings reverse in 22.4\% of
comparable cells and worst-decile-delay orderings in 23.9\%, compared with
8.6\% for robot travel time. The effect also differs across controllers.
As shown in Fig.~\ref{fig:population-shifts}, the matched deadline shift
reduces pedestrian delay under reactive control and Robot-ORCA but increases
it under early yielding. The population change therefore alters the relative
burden associated with different navigation strategies rather than producing
a common shift across controllers.

\subsection{Comparisons across Population Pairs}

The same pattern appears across all 28 population pairs.
Table~\ref{tab:main-results} pools the three controller pairs after
excluding ties. Robot travel-time orderings reverse in 8.7\% of eligible
comparisons, compared with 25.8\% for mean pedestrian delay and 24.8\% for
worst-decile delay.

Restricting the analysis to cells in which all three controllers complete
every episode under both populations reduces the reversal rates to 5.0\%,
21.5\%, and 22.1\%, respectively. The higher sensitivity of
pedestrian-burden rankings therefore remains after controller failures are
removed from the comparison.

\begin{table}[!b]
\caption{Reversals in Controller Rankings across Population Pairs}
\label{tab:main-results}
\centering
\footnotesize
\setlength{\tabcolsep}{3.2pt}
\begin{tabular}{@{}lcc@{}}
\toprule
Metric & All cells & Completed cells \\
\midrule
Robot travel time & 8.7\% & 5.0\% \\
Mean pedestrian delay & \textbf{25.8\%} & \textbf{21.5\%} \\
Worst-decile delay & \textbf{24.8\%} & \textbf{22.1\%} \\
\bottomrule
\end{tabular}
\end{table}

\subsection{Aggregate Controller Outcomes}

The controllers also exhibit distinct aggregate performance
trade-offs. Table~\ref{tab:controller-outcomes} summarizes outcomes over the
full matched grid. Reactive control achieves the highest completion rate at
90.3\%, while Robot-ORCA produces the lowest mean pedestrian delay (0.65~s)
and worst-decile delay (2.80~s). Early yielding has the lowest mean delay among
the two non-reciprocal controllers but also the highest timeout rate.

\begin{table}[!t]
\caption{Controller Outcomes over the Full Matched Grid}
\label{tab:controller-outcomes}
\centering
\footnotesize
\setlength{\tabcolsep}{2.5pt}
\begin{tabular}{@{}lrrr@{}}
\toprule
Outcome & Reactive & Early yielding & Robot-ORCA \\
\midrule
Success (\%) & 90.3 & 81.5 & 87.8 \\
Pedestrian collision (\%) & 2.6 & 5.4 & 2.1 \\
World collision (\%) & 0.1 & 0.0 & 7.1 \\
Timeout (\%) & 7.0 & 13.0 & 3.1 \\
Robot travel time (s) & 22.9 & 28.4 & 24.2 \\
Minimum clearance (m) & 0.70 & 0.70 & 0.73 \\
Mean added delay (s) & 1.25 & 1.06 & 0.65 \\
Worst-decile delay (s) & 5.13 & 4.42 & 2.80 \\
\bottomrule
\end{tabular}
\end{table}

Robot-ORCA performs best on aggregate pedestrian delay, whereas reactive
control completes more runs. Most Robot-ORCA collisions are contacts with
the environment rather than pedestrians, accounting for the difference
between its pedestrian- and world-collision rates in
Table~\ref{tab:controller-outcomes}.

\subsection{Spatial Constraint and Tail Burden}

Pedestrian burden is concentrated in the tail. Across the controllers,
worst-decile pedestrian delay is approximately four times the corresponding
population mean. Population shifts can also affect the tail differently
from the mean, as illustrated by the early yielding response to the matched
time-pressure intervention in Fig.~\ref{fig:population-shifts}.

\begin{figure*}[t]
    \centering
    \includegraphics[width=0.88\textwidth]{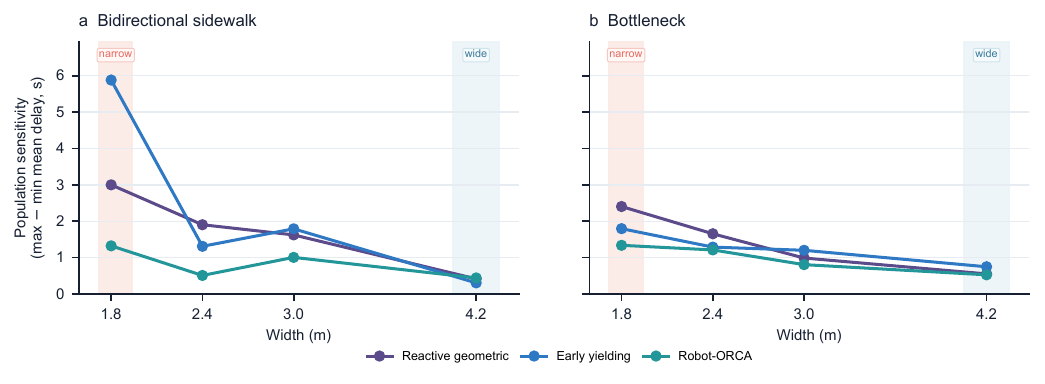}
    \caption{Population sensitivity across available widths.
    Population sensitivity is the range of mean pedestrian delay across
    population conditions for a fixed controller and physical cell, shown for
    (a) bidirectional sidewalks and (b) bottlenecks.}
    \label{fig:width}
\end{figure*}

Available space strongly affects the magnitude of population sensitivity.
As shown in Fig.~\ref{fig:width}, the range of mean pedestrian delay across
populations is larger on narrow sidewalks and bottlenecks. Pooled population
sensitivity decreases from 2.23~s at 1.8~m width to 0.50~s at 4.2~m, with the
same overall direction for each controller. When analysis is restricted to
cells completed by all controllers, width is no longer monotonic, indicating
that part of the stronger narrow-space effect coincides with controller
failures.

\subsection{Mapping and Profile Structure}

The population effect persists when the mapping from responses to motion is
changed. Table~\ref{tab:mapping} compares the four mappings on the representative
subset. Mappings B and C preserve more than 91\% of Mapping A's mean- and
tail-delay controller orderings and produce similar reversal rates. The
simpler direct Mapping D yields higher reversal rates, reaching 37.5\% for
mean delay and 35.8\% for worst-decile delay.

\begin{table}[!t]
\caption{Ranking Reversals and Agreement Across Mappings from Responses to Motion}
\label{tab:mapping}
\centering
\footnotesize
\setlength{\tabcolsep}{2.6pt}
\begin{tabular}{@{}lrrrr@{}}
\toprule
& \multicolumn{2}{c}{Reversal rate} & \multicolumn{2}{c}{Agreement with A} \\
\cmidrule(lr){2-3}\cmidrule(l){4-5}
Mapping & Mean & Tail & Mean & Tail \\
\midrule
A: original & 24.5\% & 26.0\% & -- & -- \\
B: ordered & 24.2\% & 28.6\% & 92.3\% & 92.6\% \\
C: equal weights & 24.6\% & 25.7\% & 92.2\% & 91.8\% \\
D: direct & 37.5\% & 35.8\% & 76.8\% & 71.0\% \\
\bottomrule
\end{tabular}
\end{table}

Changing the association structure among persona profiles also changes the
magnitude of the effect. Across five shared populations, randomly reassigning
complete motion profiles among persona identities reduces mean/tail reversal
rates from 28.4\%/26.9\% to 19.8\%/18.8\%. Independently shuffling each
parameter column gives 26.8\%/29.7\%. Thus, shifted parameter distributions
continue to produce ranking changes after profile associations are disrupted,
while the associations themselves affect how frequently reversals occur. In
the single-block analyses, time pressure and caution produce more ordering
changes than robot trust alone.

\subsection{Pedestrian-Dynamics and LLM Robustness}

The qualitative difference between pedestrian burden and robot travel time
also appears under a second pedestrian dynamics model. On cells completed by all
three controllers under both backends, Ped-ORCA produces mean- and
worst-decile-delay reversal rates of 10.5\% and 9.8\%, compared with 3.2\% for
robot travel time. Under the Human-like backend, the corresponding rates are
35.2\%, 34.6\%, and 2.4\%. The two dynamics models differ substantially in
reversal magnitude, while both produce more pedestrian-burden ranking reversals
than robot travel-time reversals.

Repeating the matched intervention with Gemini 3.5 Flash Lite gives the same
qualitative pattern. With the same persona IDs, reversal rates are
25.6\% for mean pedestrian delay, 26.5\% for worst-decile delay, and 9.1\%
for robot travel time. Probe-defined cohort membership can change across
language models, so this comparison is restricted to the matched intervention.

\section{Discussion}

\subsection{Population-Conditioned Evaluation}

The observed ranking reversals indicate that comparative evaluations of social-navigation strategies can depend on the behavioral population used for evaluation. With physical episodes and navigation strategies held fixed, changes in pedestrian motion profiles frequently altered strategy rankings based on pedestrian burden. This extends prior work on human variation in social navigation by showing that behavioral diversity can affect not only the simulated interaction itself, but also the comparative evaluation of navigation strategies.  

The stronger sensitivity of pedestrian burden suggests that population shift can affect how interaction costs are distributed between the robot and nearby pedestrians, even when robot-centered performance remains comparatively stable. Variation in speed, clearance, yielding, reaction, detouring, and gap acceptance can interact differently with navigation strategies, so a strategy comparison that appears stable in robot travel time may still vary substantially in pedestrian burden across populations. These patterns suggest population-specific design considerations, including clearer passage negotiation under time pressure, less forced slowing or detouring for mobility-support profiles, and greater usable clearance or earlier, more predictable conflict resolution for cautious or low-trust profiles. Such differences may become more consequential in narrow sidewalks and bottlenecks, where limited spatial flexibility leaves less room to absorb behavioral variation.  

\subsection{Implications for Social Navigation Evaluation}

The population dependence observed here suggests that behavioral population composition should be treated as an evaluation dimension alongside geometry, density, and other scenario conditions. A benchmark can include heterogeneous pedestrians while still evaluating every navigation strategy under a single fixed mixture of behaviors; such a setup may obscure population-specific changes in relative performance. Matched population replay provides a direct way to assess this dependence by evaluating the same physical episodes across alternative pedestrian populations and reporting robot-centered outcomes together with mean and tail pedestrian burden. Population robustness therefore provides an additional dimension for social-navigation evaluation.

To extend population-robustness evaluation beyond controlled synthetic populations, the same framework could be extended to empirically measured populations. Future datasets that combine contextual attributes with observed human--robot motion could provide grounded distributions of clearance, yielding, speed, and gap acceptance for different deployment settings, allowing population robustness to be evaluated using observed rather than synthetic behavioral variation.

\section{Limitations}
The behavioral populations in this study are synthetic and should be interpreted as controlled simulation inputs rather than estimates of particular cities, cultures, or demographic groups. The persona-to-motion pipeline is not calibrated using joint observations of persona attributes, encounter responses, and pedestrian motion from the same participants. The matched time-pressure intervention is similarly a controlled perturbation of the synthetic records and should not be interpreted as an estimate of the causal effect of deadlines on real pedestrians. Alternative mappings, pedestrian dynamics models, and language models preserve the main qualitative pattern while changing reversal magnitudes and some controller orderings. The simulation also abstracts several aspects of real pedestrian--robot interaction. Pedestrians are represented as disk agents, assistive-device geometry and group behavior are not modeled, and the robot receives perfect local state information. The Human-like dynamics analysis is limited to a smaller obstacle-free subset. Future work can combine empirically measured behavioral populations with richer pedestrian models and real-world robot experiments.

\section{Conclusion}
This paper introduced PopNavShift to test the stability of social-navigation strategy evaluations across pedestrian population shifts. Across matched episodes, rankings based on pedestrian burden were more sensitive than those based on robot travel time, particularly in spatially constrained settings. Changing only time pressure for the same personas frequently reversed rankings based on mean and worst-decile pedestrian delay, and this pattern persisted across alternative mappings, pedestrian dynamics models, and language models. These results support evaluating social-navigation strategies across multiple behavioral populations and reporting pedestrian burden alongside robot-centered performance.

\section*{Acknowledgment}

OpenAI GPT assisted with experimental code (Sections~III--V) and language
editing. Google Gemini generated the robot encounter responses used in
Sections~III--V.
The complete implementation and exact response-to-motion formulas will be
released upon publication.

\smallskip
\noindent{\small\sffamily Project webpage: \href{https://tantansir.github.io/PopNavShift/}{\textcolor{projectlink}{tantansir.github.io/PopNavShift}}}\par

\emergencystretch=2em
\sloppy
\bibliographystyle{IEEEtran}
\bibliography{references}

\end{document}